\documentclass{article}
\usepackage{microtype}
\usepackage{graphicx}
\usepackage{subfigure}
\usepackage{booktabs} 

\usepackage{hyperref}

\usepackage[accepted]{icml2023}

\usepackage{amsmath}
\usepackage{amssymb}
\usepackage{mathtools}
\usepackage{amsthm}
\usepackage{arydshln}

\usepackage[capitalize,noabbrev]{cleveref}

\theoremstyle{plain}
\newtheorem{theorem}{Theorem}[section]

\newtheorem{lemma}[theorem]{Lemma}
\newtheorem{corollary}[theorem]{Corollary}
\theoremstyle{definition}
\newtheorem{definition}[theorem]{Definition}

\theoremstyle{remark}

\usepackage[textsize=tiny]{todonotes}

\newcommand{\changes}[1]{{#1}}

\graphicspath{{figures/}{}}

\icmltitlerunning{Minimum Width of Leaky-ReLU Neural Networks for Uniform Universal Approximation}

\begin{document}

\twocolumn[
\icmltitle{Minimum Width of Leaky-ReLU Neural Networks\\ for Uniform Universal Approximation}
\icmlsetsymbol{equal}{*}

\begin{icmlauthorlist}
\icmlauthor{Li'ang Li}{equal,bnu}
\icmlauthor{Yifei Duan}{equal,bnu}
\icmlauthor{Guanghua Ji}{bnu}
\icmlauthor{Yongqiang Cai}{bnu}
\end{icmlauthorlist}

\icmlaffiliation{bnu}{School of Mathematical Sciences, Laboratory of Mathematics
and Complex Systems, MOE, Beijing Normal University, Beijing,
100875, China}

\icmlcorrespondingauthor{Yongqiang Cai}{caiyq.math@bnu.edu.cn}
\icmlkeywords{Machine Learning, ICML}

\vskip 0.3in
]

\printAffiliationsAndNotice{\icmlEqualContribution} 

\begin{abstract}

    The study of universal approximation properties (UAP) for neural networks (NN) has a long history. When the network width is unlimited, only a single hidden layer is sufficient for UAP. In contrast, when the depth is unlimited, the width for UAP needs to be not less than the critical width $w^*_{\min}=\max(d_x,d_y)$, where $d_x$ and $d_y$ are the dimensions of the input and output, respectively. Recently, \cite{cai2022achieve} shows that a leaky-ReLU NN with this critical width can achieve UAP for $L^p$ functions on a compact domain $\mathcal{K}$, \emph{i.e.,} the UAP for $L^p(\mathcal{K},\mathbb{R}^{d_y})$. This paper examines a uniform UAP for the function class $C(\mathcal{K},\mathbb{R}^{d_y})$ and gives the exact minimum width of the leaky-ReLU NN as  $w_{\min}=\max(d_x,d_y)+\Delta (d_x, d_y)$, where $\Delta (d_x, d_y)$ is the additional dimensions for approximating continuous functions with diffeomorphisms via embedding. To obtain this result, we propose a novel lift-flow-discretization approach that shows that the uniform UAP has a deep connection with topological theory.

\end{abstract}

\section{Introduction}
    The universal approximation theorem is important for the development of artificial neural networks. Artificial neural networks can approximate functions with arbitrary precision, this fact reveals the great potential of neural networks and provides important guarantees for their development. \cite{cybenko1989approximation} produces the original universal approximation theorem, stating that an arbitrarily wide feedforward neural network with a single hidden layer and sigmoid activation function can arbitrarily approximate continuous function. \cite{hornik1991approximation} later demonstrated that the key to the universal approximation property lies in the multilayer and neuron architecture rather than the choice of an activation function. Then, \cite{Leshno1993Multilayer} show that for a continuous function $f:\mathcal{K} \rightarrow R^{d_y}$ defined on a compact set $\mathcal{K} \subseteq R^{d_x}$ can be approximated by a single hidden layer neural network, if and only if, the activation function is a nonpolynomial function.

    After solving the activation function's theoretical problem, the field of vision naturally shifted to a consideration of the width and depth of the neural network. With the gradual development of deep neural networks, researchers have begun to pay attention to how to theoretically analyze the expressiveness of networks. \cite{daniely2017depth} simplifies the proof that the expressive ability of the three-layer neural network is superior to that of the two-layer neural network. For any positive integer $k$, \cite{telgarsky2016benefits} shows that there are neural networks with $\Theta(k^3)$ layers and fixed widths that cannot be approximated by networks with $\mathcal{O}(k)$ layers unless they have $\Omega(2^k)$ nodes \footnote{$\Theta(k^3)$ means that it is bound both above and below by $k^3$ asymptotically; $\mathcal{O}(k)$ means that it is bounded above by $k$ asymptotically; $\Omega(2^k)$ means that it is bounded below by $2^k$ asymptotically. }. The universal approximation theorem explains that deep-bounded neural networks with suitable activation functions are universal approximators. \cite{Lu2017Expressive} explained that a neural network with a bounded width can also be a universal approximator, such as the width-$(d_x+4)$ ReLU networks, where $d_x$ is the input dimension. \cite{Lu2017Expressive} also shows that a ReLU network of width $d_x$ cannot be used for universal approximation.
    
    \begin{table*}\label{tab:main}
    \centering
    \caption{A summary of the known minimum width of feed-forward neural networks for universal approximation. $^\dagger$ }
    \setlength{\tabcolsep}{1mm}{
        \begin{tabular}{cccc}
            \toprule[1.5pt]
            References & Functions & Activation & Minimum width\\
            \hline
            \cite{Hanin2018Approximating} & $\mathcal{C}(\mathcal{K},\mathbb{R})$ & ReLU & $w_{\min}=d_x+1$\\
            \hline
            \cite{Park2021Minimum} & $L^p(\mathbb{R}^{d_x},\mathbb{R}^{d_y})$ & ReLU & $w_{\min}=\max(d_x + 1 ,d_y)$\\
            \cite{Park2021Minimum} & $\mathcal{C}([0,1],\mathbb{R}^2)$ & ReLU & $w_{\min}=3$\\
            \cite{Park2021Minimum} & $\mathcal{C}(\mathcal{K},\mathbb{R}^{d_y})$ & ReLU + STEP & $w_{\min}=\max(d_x + 1,d_y)$\\
            \hline
            \cite{cai2022achieve} & $L^p(\mathcal{K},\mathbb{R}^{d_y})$ &Leaky-ReLU & $w_{\min} = \max(d_x,d_y,2)$\\
            \cite{cai2022achieve} & $\mathcal{C}(\mathcal{K},\mathbb{R}^{d_y})$ &ReLU + FLOOR & $w_{\min} = \max(d_x,d_y,2)$\\
            \hline
            Ours (Theorem \ref{th:main_min_width}) & $\mathcal{C}(\mathcal{K},\mathbb{R}^{d_y})$ &Leaky-ReLU &$w_{\min}=\max(d_x,d_y)+\Delta (d_x , d_y)$\\
            Ours (Lemma \ref{th:w_upper_bound}) & $\mathcal{C}(\mathcal{K},\mathbb{R})$ &Leaky-ReLU &$w_{\min}=d_x+1$\\
            \bottomrule[1.5pt]
        \end{tabular}\\
        $\dagger$ {$d_x$ and $d_y$ are the input and output dimensions, respectively. $\mathcal{K} \subset \mathbb{R}^{d_x}$ is a compact domain and $p \in [1,\infty)$. }
        }
    \end{table*}
    
    Many studies, such as \cite{Beise2020Expressiveness, Hanin2018Approximating, Park2021Minimum}, have shown that for a narrow neural network (the width is not greater than the input dimension), it is difficult to attain the UAP. \cite{Nguyen2018Neural} noted that deep neural networks with a specific type of activation function generally need to have a width larger than the input dimension to guarantee that the network can produce disconnected decision regions. For ReLU networks, \cite{Park2021Minimum} proved that the minimum width for $L^p$-UAP is $w_{\min} = \max (d_x+1,d_y)$ and summarized the known upper/lower bounds on the minimum width for universal approximation. Furthermore, conclusions related to the UAP of continuous functions have yet to be studied. \cite{Park2021Minimum} and \cite{cai2022achieve} demonstrate the minimum width of some neural networks for $C$-UAP using noncontinuous activation functions. If only continuous monotonically increasing activation functions are used, the known minimum width is restricted to the ReLU NN for function class $\mathcal{C}([0,1],\mathbb{R}^2)$, where the critical width is $w_{\min} = 3$. Table \ref{tab:main} provides a summary of the known minimum width for UAP.

    To determine the minimum width of uniform UAP on $\mathcal{C}(\mathcal{K},\mathbb{R}^{d_y})$, we introduce a novel scheme called \emph{lift-flow-discretization approach}. Based on the close relationship between uniform UAP and topology, the functions are embedded in high-dimensional diffeomorphisms or flow maps, and feedforward neural networks are used to approximate these flow maps. Finally, we determine the minimum width of leaky-ReLU neural networks for $C$-UAP on $\mathcal{C}(\mathcal{K},\mathbb{R}^{d_y})$ to be $w_{\min}=\max(d_x,d_y)+\Delta (d_x, d_y)$.

    \subsection{Contributions}
    \begin{itemize}
        \item[1.] Theorem \ref{th:main_min_width} states that the minimum width of leaky-ReLU networks for $\mathcal{C}(\mathcal{K},\mathbb{R}^{d_y})$ can be expressed as $w_{\min}=\max(d_x,d_y)+\Delta (d_x , d_y)$. This is the first time that the minimum width for the universal approximation of leaky-ReLU networks is systemly studied by topology. It is worth mentioning that the previous results for the minimum width for the uniform approximation are based on discontinuous activation functions. The conclusion of this paper is based on continuous activation functions such as the leaky-ReLU function.
        
        \item[2.] Section \ref{section3} presents a novel approach for approximating continuous functions using a feedforward neural network from the perspective of topology. The lift-flow-discretization approach of combining topology and neural network approximation is the key to the proof in this paper. Our approach is generic for strictly monotone continuous activations, as they all correspond to diffeomorphisms.
    \end{itemize}

    \subsection{Related work}
    
    \textbf{Width and depth bounds.}
    Theoretical analyses of the expressive power of neural networks have taken place over the years. \cite{cybenko1989approximation} proposed a prototype of the early classic universal approximation theorem. Continuous univariate functions over bounded domains can be fitted with arbitrary precision using the sigmoid activation function. \cite{Hornik1989Multilayer, Leshno1993Multilayer, barron1994approximation} obtained similar conclusions and extended them to a large class of activation functions, revealing the relationship between universal approximation and network structure.
    
    The effect of neural network width on expressiveness is an enduring question. \cite{Sutskever2008Deep, LeRoux2008Representational} and \cite{Montufar2014Universal} reveal the impact of depth and width, especially width, on the general approximation of belief networks, and networks with too narrow a width cannot complete the approximation task. The width has important research value for many emerging networks and different activation functions. Conventional conclusions tell us that networks with appropriate activation functions under bounded depths are universal approximators. Correspondingly, \cite{Lu2017Expressive} proposed a general approximation theorem for ReLU networks with bounded widths. \cite{Hanin2018Approximating} also studied in the ReLU network, whose input dimension is $d_{x}$,  hidden layer width is at most $w$ and depth is not limited. To fit any continuous real-valued function, the minimum value of $w$ is exactly $d_{x}+1$.
    
    For a deep neural network that satisfies the activation function $\sigma(\mathbb{R})=\mathbb{R}$, to learn the disconnected regions, it is usually necessary to make the network width larger than the input dimension. If the network is narrow, the paths connecting the disconnected regions yield high-confidence predictions \cite{Nguyen2018Neural}. \cite{Chong2020closer} gives a direct algebraic proof of the universal approximation theorem, and \cite{Beise2021decision} reveals the fundamental reason why the universal approximation of network functions with width $w\leq d_x$ from $\mathbb{R}^{d_x}$ to $\mathbb{R}$  is impossible.
    
    \cite{Park2021Minimum} gives the first definitive results for the critical width enabling the universal approximation of width-bounded networks. The minimum width for the $L^p$ functions is $\max(d_x + 1, d_y)$ using the ReLU activation functions. \cite{Park2021Minimum} also shows that this conclusion is unsuitable for the uniform approximation of the ReLU network, but it still holds using the ReLU+STEP activation function. \cite{cai2022achieve} shows that minimum widths for the $C$-UAP and $L^p$-UAP 
    on compact domains have a universal lower bound $w_{\min} = \max(d_x, d_y)$. \cite{cai2022achieve} also shows the minimum width for the uniform approximation with some additional threshold activation functions.
    
    \textbf{Homeomorphism properties of networks.}
    Residual networks (ResNets) are an advanced deep-learning architecture for supervised learning problems. \cite{rousseau2018residual} shows that a continuous flow of diffeomorphisms governed by ordinary differential equations can be numerically implemented using the mapping component of ResNets.
    
    Neural ordinary differential equations (neural ODEs) turn the neural network training problem into a problem of solving differential equations and can make the discrete ResNet continuous. As a deep learning method, \cite{Teshima2020Universal} shows the universality of discrete neural ODEs with the condition that the source vectors $f_i(z) \in \mathcal{H}$, where $\mathcal{H}$ is a universal approximator for the Lipschitz functions. \cite{Ruiz-Balet2021Neural} provide $L^2$-UAP of neural ODE $\dot x = W \sigma(Ax+b)$ for $L^2(\mathcal{K},\mathbb{R}^d)$. \cite{Zhang2019Approximation} shows that neural ODEs with extra dimensions are universal approximators for homeomorphisms.

    Invertible neural networks have diffeomorphic properties, and many flow models can also be used as universal approximators. \cite{Huang2018Neural} shows that neural autoregressive flows are universal approximators for continuous probability distributions. \cite{Teshima2020Couplingbased} indicates that normalizing flow models based on affine coupling also have UAP. \cite{Kong2021Universal} 
    shows
    that residual flows are universal approximators in maximum mean discrepancies.

    \subsection{Organization}
    
    We first define the necessary notation and the main results and give the proof ideas in Section \ref{section2}. In Section \ref{section3}, we present our \emph{lift-flow-discretization approach} and complete the proof of Theorem \ref{th:main_min_width}. In section \ref{sec4} we give examples of analysis in special cases by considering the influence of output dimensions. In Section \ref{sec:discussion}, we give an outlook on the direction of our current work. All formal proofs are provided in the appendix.

    \section{Main results}
    \label{section2}

    We consider the standard feedforward neural network with the same number of neurons at each hidden layer. We say a $\sigma$-NN with depth $L$ is a function with inputs $x\in \mathbb{R}^{d_x}$ and outputs $y\in\mathbb{R}^{d_y}$, which has the following form:
    \begin{align} \label{eq:FNN}
        y &= f_{NN,L}(x)= y_L  \\
        & = W_{L+1} \sigma \left(W_L\left( \cdots \sigma \left( W_1 x + b_1\right) + \cdots\right)
        + b_L\right) + b_{L+1},\nonumber
    \end{align}
    where $b_i$ are vectors, $W_i$ are matrices and $\sigma(\cdot)$ is the activation function. We mainly consider the number of neurons in all the layers to be the same $N$.
    In this case, $W_i \in \mathbb{R}^{N\times N}, b_i \in \mathbb{R}^{N}, i\in \{1,\cdots,L + 1\} $, except $W_{1} \in \mathbb{R}^{N\times d_x}$, $W_{L+1} \in \mathbb{R}^{d_y \times N}$ and $b_{L+1} \in \mathbb{R}^{d_y}$. 
    We denote the set of all networks in Eq.~(\ref{eq:FNN}) as $\mathcal{N}_{N,L}(\sigma)$, and $\mathcal{N}_{N}(\sigma) = \bigcup_L \mathcal{N}_{N,L}(\sigma)$. 
    The activation function is crucial for the approximation power of the neural network. Our main results are for the following leaky-ReLU activations function with a fixed parameter $\alpha \in \mathbb{R}^+\setminus\{1\}$,
    \begin{align}
        \sigma(x) = \sigma_\alpha(x) = 
        \begin{cases}
            x, & x>0,\\
            \alpha x, & x\le 0.\\
        \end{cases}
    \end{align}
    
    \subsection{Main theorem}
    
    Our main theorem is the following Theorem \ref{th:main_min_width}, which provides the exact minimum width of the leaky-ReLU networks that process uniform universal approximations.
    \begin{definition}
        We say the leaky-ReLU networks with width $N$ have $C$-UAP or $L^p$-UAP if the set $\mathcal{N}_{N}(\sigma)$ is dense in $C(\mathcal{K},\mathbb{R}^{d_y})$ or $L^p(\mathcal{K},\mathbb{R}^{d_y})$, respectively. 
    \end{definition}
    
    \begin{theorem}
        \label{th:main_min_width}
        Let $\mathcal{K} \subset \mathbb{R}^{d_x}$ be a compact set; then, for the continuous function class $C(\mathcal{K},\mathbb{R}^{d_y})$, the minimum width $w_{\min}$ of leaky-ReLU neural networks having $C$-UAP is $w_{\min}=\max(d_x,d_y)+\Delta(d_x, d_y)$, where $\Delta(d_x,d_y)$ is the auxiliary for approximating continuous functions with diffeomorphisms via embedding. Thus, $\mathcal{N}_{N}(\sigma)$ is dense in $C(\mathcal{K},\mathbb{R}^{d_y})$ if and only if $N \ge w_{\min}$.
    \end{theorem}
    
    \begin{figure*}[htp!]
        \center
        \includegraphics[width=16cm]{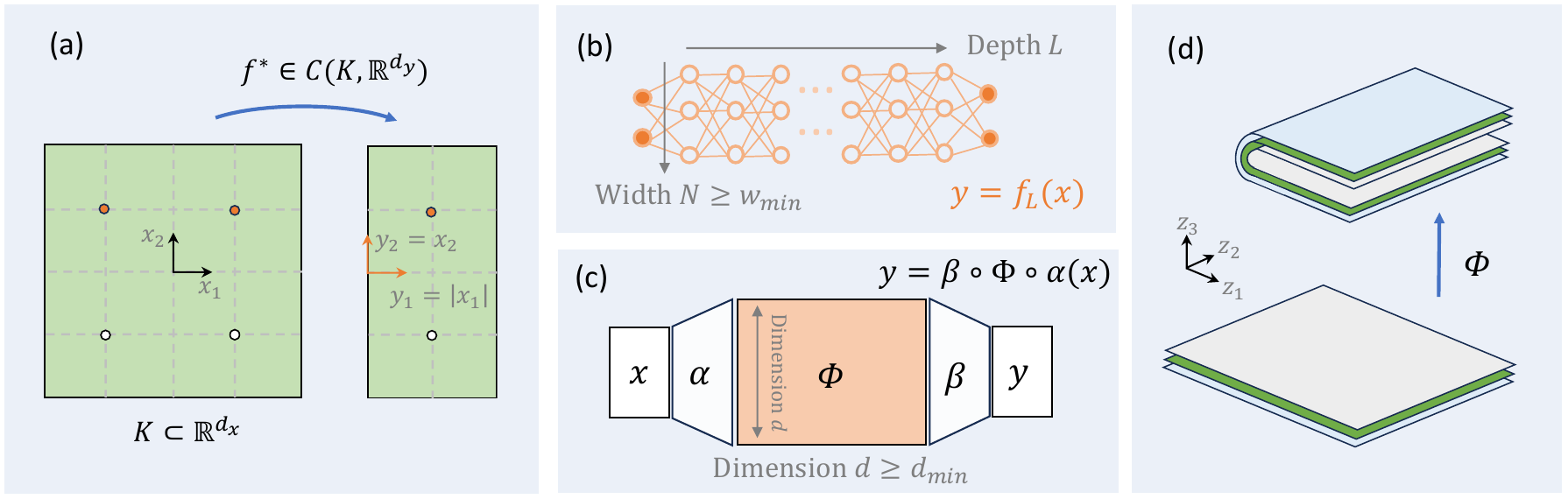}
        \caption{Continuous function, neural network, and diffeomorphism.
        (a) An example function from $\mathcal{K} \subset \mathbb{R}^{d_x}$ to $\mathbb{R}^{d_y}$.
        (b) Feedforward neural networks with depth $L$ and width $N$.
        (c) Approximate $f^*$ using two linear transformations, $\alpha$ and $\beta$, and a diffeomorphism $\Phi$.
        (d) An intuitive construction of $\Phi$, where $\alpha$ and $\beta$ are a lift and a projection, respectively, \emph{i.e.}, $\alpha(x_1,x_2)=(x_1,x_2,0), \beta(z_1,z_2,z_3)=(z_1,z_2)$. }
        \label{fig:main}
    \end{figure*}

    Before giving the proof, let's emphasize the points of Theorem~\ref{th:main_min_width}. First, if the width $N$ of the leaky-ReLU networks is smaller than $w_{\min}$, then there is a continuous function $f^* \in C(\mathcal{K},\mathbb{R}^{d_y})$ that cannot be well approximated, \emph{i.e.} there is a positive constant $\varepsilon>0$ such that $\|f-f^*\| > \varepsilon$ for all $f \in \mathcal{N}_N(\sigma)$. For the case of $\mathcal{K}=[-1,1]^{d_x},d_y=1$, the function $f^*$ can be chosen as $f^*(x) = \|x\|^2$. The reason can be given by considering the topological properties of the level sets \cite{Johnson2019Deep,cai2022achieve}.
    
    Second, if $N = w_{\min}$, then for any $f^* \in C(\mathcal{K},\mathbb{R}^{d_y})$ and any $\varepsilon>0$, we can construct a leaky-ReLU network $f_L$ with width $N$ and depth $L$ such that $\|f-f^*\| < \varepsilon$. We will introduce the construction scheme later. 
    
    Lastly, the formula of $w_{\min}$ includes a special function $\Delta (d_x,d_y)$ which is defined from topology. It is well known to topologists that any $f^* \in \mathcal{C}(\mathcal{K},\mathbb{R}^{d_y})$ can be well approximated by embeddings provided the dimensional $d_y$ is large (see Page~26 of \cite{Hirsch1976Differential} for example). 
    As a consequence, there is a dimension $d$ large enough, such that $f^*$ can be approximated by functions of the form $\beta \circ \Phi \circ \alpha$ where $\Phi$ is a diffeomorphism of $\mathbb{R}^d$, $\alpha$ and $\beta$ are two linear maps. Figure.~\ref{fig:main} gives an example of such approximations. Denote the minimum of such dimensions as $d_{\min}$, which is necessarily larger than $\max(d_x,d_y)$, then  $\Delta (d_x,d_y)$ is defined as $\Delta (d_x,d_y) = d_{\min} - \max(d_x,d_y)$, the auxiliary for approximating continuous functions with diffeomorphisms via embedding.
    
    We emphasize that determining the value of $d_{\min}$ or $\Delta (d_x,d_y)$ is a purely topological problem and the main message of our Theorem~\ref{th:main_min_width} is that $d_{\min}=w_{\min}$. In other words, the minimum width of leaky-ReLU networks is equivalent to the minimum dimension for embedding. The following Lemma \ref{th:w_upper_bound} shows a simple application of this connection, which implies $w_{\min}=d_x+1$ when $d_y=1$.
    \begin{lemma}
        \label{th:w_upper_bound} 
        For any continuous function $f^* \in C(\mathcal{K},\mathbb{R})$ on compact domain $\mathcal{K} \subset \mathbb{R}^{d_x}$, and $\varepsilon>0$, there is a leaky-ReLU network $f_L(x)$ with depth $L$ and width $d+1$ such that $\|f_L(x) - f^*(x)\| \le \varepsilon$ for all $x$ in $\mathcal{K}$. In addition, $d_x+1$ is the minimal width.
    \end{lemma}
    The proof of Lemma \ref{th:w_upper_bound} will be given in Section \ref{section3}, where the network $f_L$ and its corresponding embedding for approximation are easy to construct. 
    % }
    
    \subsection{Proof ideas}
    
    Now, we provide the proof scheme which will be detailed in the next section. Theorem~\ref{th:main_min_width} can be split into two parts: 
    \begin{enumerate}
        \item $\max(d_x,d_y)+\Delta(d_x, d_y)$ is a lower bound, \emph{i.e.}, $w_{\min} \ge d_{\min}$,
        \item $\max(d_x,d_y)+\Delta(d_x, d_y)$ is also an upper bound, \emph{i.e.}, $w_{\min} \le d_{\min}$.
    \end{enumerate}
    
    Part 1 is formally stated as the following lemma.
    \begin{lemma}\label{2.3}
     For any continuous function $f^* \in C(\mathcal{K},\mathbb{R}^{d_y})$ on compact domain $\mathcal{K} \subset \mathbb{R}^{d_x}$, if $f^*$ can be approximated by the leaky-ReLU network $f_L$ with width $w_{\min}$ and depth $L$, then $f^*$ can be approximated by $\beta \circ \Phi \circ \alpha$, where $\Phi$ is a diffeomorphism of $\mathbb{R}^{w_{\min}}$, $\alpha$ and $\beta$ are two linear maps.
    
    \end{lemma}

    The lemma can be proved constructively. Note that the leaky-ReLU network $f_L$ is already a function of form $\beta \circ \tilde \Phi \circ \alpha$, except the inner function $\tilde \Phi$ is not necessarily a diffeomorphism. However, $\tilde \Phi$ is a composition of maps in $\mathbb{R}^{w_{\min}}$ which can be slightly modified to be diffeomorphisms. The square matrix $W_i$ can be well approximated by some nonsingular matrix $\hat W_i$ and the activation $\sigma(x)$ can be well approximated by its smooth version $\hat \sigma(x)$. The map $\hat W_i x + b_i$ and $\hat\sigma(x)$ are diffeomorphisms. Hence replaced $W_i$ and $\sigma$ by $\hat W_i$ and $\hat\sigma$, respectively, the network $f_L$ becomes the expected form $\beta \circ \Phi \circ \alpha$ where $\Phi$ is a diffeomorphism. 

    As the inverse of Part 1, Part 2 of Theorem~\ref{th:main_min_width} can be also proved constructively. The aim is to construct leaky-ReLU networks with width $w=d_{\min}$ to approximate diffeomorphisms of $\mathbb{R^{d_{\min}}}$. In the next section, we will give such a construction which is named the lift-flow-discretization approach. The core idea is employing the splitting method to discretize a diffeomorphism as a composition of linear and leaky-ReLU maps.

    \section{Lift-flow-discretization approach}
    \label{section3}

    We reformulate network (\ref{eq:FNN}) as follows:
    \begin{align}
        f_L(x) = W_{L+1} \Phi_L( W_1 x + b_1) + b_{L+1},
    \end{align}
    where $\Phi_L$ is a map from $\mathbb{R}^N$ to $\mathbb{R}^N$ and $W_1 x+b_1$ and $W_{L+1} \Phi + b_{L+1}$ are linear maps. Since we use the leaky-ReLU activation and the weight matrix in 
    (\ref{eq:FNN}), it can 
    be assumed to be nonsingular, the map $\Phi_L$ is a homeomorphism. Motivated by the recent work of \cite{duan2022vanilla}, which shows that leaky-ReLU networks can approximate flow maps, we propose an approach to approximate functions $f^*$ in $C(\mathcal{K},\mathbb{R}^{d_y})$ by lifting it as a diffeomorphism $\Phi$ and then we approximate $\Phi$ by flow maps and neural networks.
    
    For any function $f^*$ in $C(\mathcal{K},\mathbb{R}^{d_y})$ and any $\varepsilon>0$, our lift-flow-discretization approach includes three steps:
    \begin{itemize}
        \item[1)](\textbf{Lift}) A lift map $\Phi \in C(\mathbb{R}^{N},\mathbb{R}^{N})$, which is an orientation preserving (OP) diffeomorphism such that
        \begin{align}\label{lift}
            \|f^*(x) - \beta \circ \Phi \circ \alpha (x)\| \le \varepsilon/3, 
            \quad
            \forall x \in \mathcal{K},
        \end{align}
        where $\alpha$ and $\beta$ are two linear maps. Without loss of generality, we can assume that the Lipschitz constants of $\alpha$ and $\beta$ are less than one. Within this notation, we say the map $\Phi$ is a lift of $f^*$. 
    
        \item[2)](\textbf{Flow}) A flow map $\phi^\tau \in C(\mathbb{R}^{N},\mathbb{R}^{N})$ corresponding to a neural ODE 
        \begin{align}\label{flow}
            z'(t) = v(z(t),t), t \in (0,\tau), \quad z(0) = x,
        \end{align}
        which satisfies $\|\Phi(x) - \phi^\tau(x)\|\le \varepsilon/3$ for all $x$ in $\alpha(\mathcal{K})$.
    
        \item[3)](\textbf{Discretization}) A discretization map $\psi \in C(\mathbb{R}^{N},\mathbb{R}^{N})$ is a leaky-ReLU network in $\mathcal{N}_N(\sigma)$ that approximates $\phi^\tau$ such that $\|\psi(x) - \phi^\tau(x)\|\le \varepsilon/3$ for all $x$ in $\alpha(\mathcal{K})$.
    \end{itemize}
    As a result, the composition $\beta \circ \psi \circ \alpha =: f_L $ is a leaky-ReLU network with width $N$, which approximates the target function $f^*$ such that
    \begin{align}\label{discretization}
        \|f^*(x) - \beta \circ \psi \circ \alpha (x)\| \le \varepsilon, 
        \quad
        \forall x \in \mathcal{K}.
    \end{align}

    \subsection{Theory of the lift-flow-discretization approach}
    
    Note that the existence of $\phi^\tau$ and $\psi$ are guaranteed by the following lemmas based on the results of \cite{caponigro2011orientation} and \cite{duan2022vanilla}. We need to construct the lift map $\Phi$, which will be constructed case by case. 
    
    \begin{lemma}\label{th:appproximate_OP_diff}
        Let $\Phi$ be an orientation preserving diffeomorphism of $\mathbb{R}^N$, $\mathcal{K}$ be a compact set in $\mathbb{R}^N$ and $\varepsilon>0$. Then, there is an ODE with tanh neural fields, whose flow map is denoted by $\phi^\tau(x_0)=x(\tau)$, 
        \begin{align}\label{eq:tanh_ODE}
          \dot{x}(t) &= v(x(t),t) 
            \\
            &\equiv 
            \sum\limits_{i=1}^M a_i(t)\tanh(w_i(t) \cdot x(t)+b_i(t)), t\in [0,\tau], \nonumber\\
            x(0) &= x_0 \in \mathbb{R}^N, 
            \quad M \in \mathbb{Z}^+,\nonumber
        \end{align}
    where $a_i,w_i \in \mathbb{R}^N$ and $b_i \in \mathbb{R}$ are piecewise constant functions of $t$, such that $\|\phi^\tau(x_0) - \Phi(x_0)\| < \varepsilon$ for all $x_0$ in $\Omega$.
    \end{lemma}
    
    \begin{figure}[htpb!]
        \centering
        \includegraphics[width=7.5cm]{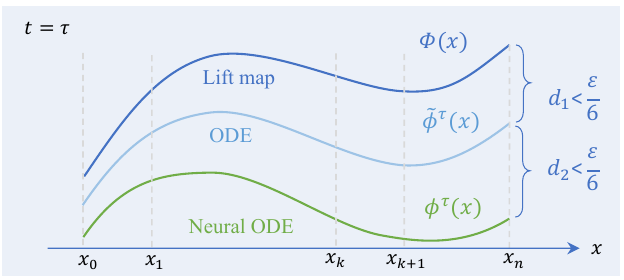}
        \caption{Sketch of the lift-flow-discretization approach. The target map ${\Phi}(x)$ is approximated by a flow map $\tilde{\phi}^\tau(x)$ of an ODE, which is further approximated by a flow map $\phi^\tau(x)$ of a neural ODE (\ref{eq:tanh_ODE}).
        }
        \label{fig:simple_proof}
    \end{figure}

    Lemma~\ref{th:appproximate_OP_diff} ensures Step 2 (`flow') of our lift-flow-discretization approach, where we
    use the flow map of a neural ODE to approximate a given orientation preserving diffeomorphism.
    
    The formal proof of the lemma can be seen in the appendix. Here, we provide the main idea of the proof.
    First, we refer to \cite{caponigro2011orientation} to prove that for any $\varepsilon>0$, there exists a flow map at the endpoint of time
    $\tilde{\phi}^\tau(x)$ of an ODE such that $\|\tilde{\phi}^\tau(x)-\Phi(x)\|<\varepsilon/2$ for all $x \in \alpha (\mathcal{K})$,
    then we use neural ODE (\ref{eq:tanh_ODE}) to approximate $\tilde{\phi}^\tau(x)$,
    there exist $(a,w,b)$ such that the flow map (denoted as $\phi^\tau(x)$) of Eq.~(\ref{eq:tanh_ODE}) satisfies
    $\|\phi^\tau(x)-\tilde{\phi}^\tau(x)\|<\varepsilon/2$, then $\|\phi^\tau(x)-\Phi(x)\|<\varepsilon$.
    
    \begin{lemma}
        \label{th:appproximate_flowmap}
        Let $\phi^\tau \in C(\mathbb{R}^{N},\mathbb{R}^{N})$ be the flow map in Lemma \ref{th:appproximate_OP_diff} and $\Omega$ be a compact set in $\mathbb{R}^N$ and $\varepsilon>0$. Then, there is a leaky-ReLU network $\psi \in \mathcal{N}_N(\sigma)$ with width $N$ and depth $L$ such that $\|\phi^\tau(x_0) - \psi(x_0)\| < \varepsilon$ for all $x_0$ in $\Omega$.
    \end{lemma}
    This lemma ensures Step 3 (discretization) of our lift-flow-discretization approach, where we find a neural network to approximate $\phi^\tau$ in Step 2.
    
    The formal proof, motivated by \cite{duan2022vanilla}, can be seen in the appendix. The main idea is to solve the ODE (\ref{eq:tanh_ODE}) by a splitting method and then approximate each split step by leaky-ReLU networks. Consider the following splitting for $v$ in (\ref{eq:tanh_ODE}),
    $ v(x,t) 
    \equiv
    \sum_{i=1}^N \sum_{j=1}^d 
    v_{ij}(x,t) 
    {e}_j$ with $v_{ij}(x,t) = a_{i}^{(j)}(t) \tanh(w_i(t) \cdot x+b_i(t)) \in \mathbb{R}$.
    Then, the flow map can be approximated by an iteration with time step $\Delta t = \tau/n$, $n \in \mathbb{Z}^+$ large enough,
    \begin{align*}
        \phi^{\tau}(x_0) \approx  x_n
        &= T_n (x_{n-1}) = T_n \circ \cdots \circ T_1 (x_0),\\
        &=
        T_n^{(N,d)} \circ \dots \circ T_n^{(1,2)} \circ T_n^{(1,1)} 
        \circ
        \cdots \circ\\
        &\quad T_1^{(N,d)} \circ \cdots \circ T_1^{(1,2)} \circ T_1^{(1,1)} 
        (x_0).
    \end{align*}
    where the $k$-th iteration is $x_{k+1} = T_k x_k = T_k^{(N,d)} \circ \dots \circ T_k^{(1,2)} \circ T_k^{(1,1)} (x_k)$,
    The map $T_k^{(i,j)}: x \to y$ in each split step is:
    \begin{align*}
        T_k^{(i,j)}:
        \left\{
        \begin{aligned} 
        & y^{(l)} = x^{(l)} , l = 1,2,..,j-1,j+1,...,d,  \\
        & y^{(j)} = x^{(j)} + \Delta t v_{ij}(x,k\Delta t).
        \end{aligned}
        \right.
    \end{align*}
    
    Combining all the approximation networks, we have $\|\phi^\tau(x_0)-\psi(x_0)\|<\varepsilon$ for all $x_0 \in \Omega$.
    
    Having reached the above conclusion, if the lift map $\Phi$ in Step 1 (lift) is constructed, we can obtain the following corollary.
    \begin{corollary}\label{th:LFD_approach}
        Let $f^* \in C(\mathcal{K},\mathbb{R}^{d}) (\mathcal{K}\subset \mathbb{R}^d)$ and $N \in \mathbb{Z}^+$. If for any $\varepsilon>0$, there is an orientation preserving diffeomorphism $\Phi$ of $\mathbb{R}^N$ and two linear maps $\alpha$ and $\beta$ such that $\|f^*(x) - \beta \circ \Phi \circ \alpha (x)\| < \varepsilon$ for all $x \in \mathcal{K}$, then there is a leaky-ReLU network $f_L \in \mathcal{N}_N(\sigma)$ with width $N$ and depth $L$ such that $\|f_L(x) - f^*(x)\| < \varepsilon$ for all $x \in \mathcal{K}$.
    \end{corollary}
    This corollary shows that $w_{\min} \le \max(d_x,d_y)+\Delta (d_x , d_y)$, which is the Part 2 of Theorem \ref{th:main_min_width}. At this point, we can use the two-step proof of Theorem \ref{th:main_min_width} to give a detailed proof of Lemma \ref{th:w_upper_bound}.

    \subsection{Proof  of Lemma \ref{th:w_upper_bound}}\label{proof:2.4}
    
    As we all know, every continuous function defined on a closed interval can be uniformly approximated as closely as desired by a polynomial function according to the Weierstrass approximation theorem.   
    Therefore, without loss of generality, we assume $f^*$ to be a polynomial function.
    Taking two linear maps $\alpha : \mathbb{R}^{d_x} \rightarrow  \mathbb{R}^{d_x +1}, (x_1,\cdots ,x_{d_x})  \rightarrow (x_1,\cdots ,x_{d_x},0)$ and $\beta :\mathbb{R}^{d_x +1} \rightarrow  \mathbb{R}, (x_1,\cdots ,x_{d_x +1})  \rightarrow x_{d_x+1}$, we can construct the mapping $\Phi : \mathbb{R}^{d_x + 1} \rightarrow  \mathbb{R}^{d_x + 1}, (x_1,\cdots ,x_{d_x+1})  \rightarrow (x_1,\cdots,x_{d_x},f^* + x_{d_x +1})$  to satisfy $f^* = \beta \circ \Phi \circ \alpha (x)$. Since $f^*$  is a continuous function, $\Phi$ is obviously a continuous one-to-one mapping. The Jacobian matrix
    $$
    \nabla \Phi  = \begin{pmatrix} 1 & 0 & \ldots  & 0 & \frac{\partial f^*}{\partial x_1} \\ 0 & 1 & \ddots & \vdots & \frac{\partial f^*}{\partial x_2} \\ \vdots & \ddots & \ddots & 0 & \vdots \\ 0 & \ldots & 0 & 1 & \frac{\partial f^*}{\partial x_{d_x}} \\ 0 & \cdots & 0 & 0 & 1 \end{pmatrix}
    $$
    is an upper triangular matrix whose determinant is one. It implies that $\Phi$ is a diffeomorphism. According to the Corollary \ref{th:LFD_approach},  leaky-ReLU network $f_L \in \mathcal{N}_{d_x +1}(\sigma)$ that satisfies the hypothesis exists, so $w_{\min} \le d_x + 1$. On the other hand, there exist continuous functions that cannot be approximated by diffeomorphisms of $\mathbb{R}^{d_x}$. For example, consider the function $f^* = \|x\|^2, x \in [-1,1]^{d_x}$. The original point $O$ is an inner point in the domain, and its image is the minimum value of the function. This means that $f^*$ maps an inner point to a boundary point, which is not possible for homeomorphisms. According to Lemma \ref{2.3}, $w_{\min} \ge d_x+1$. Combining the above two parts, the Lemma is proved.

    Our 'lift-flow-discretization' approach deeply connects the minimal width to topology theory, providing that the activation is a one-dimensional diffeomorphism.
    
    \section{Effect of the output dimension}
    \label{sec4}
    % \label{sec:output_dimension}
    
    Now we give the practical application of $\Delta(d_x,d_y)$ for minimum width analysis which considers the case of $d_y \ge d_x+1$. We examine the approximation power of leaky-ReLU networks with width $N=d_y$.
    
    We emphasize the homeomorphism properties. Leaky-ReLU, the nonsingular linear transformer, and their inverse are continuous and homeomorphic. Since compositions of homeomorphism are also homeomorphism, we have the input-output map as a homeomorphism. Note that a singular matrix can be approximated by nonsingular matrices, therefore we can restrict the weight matrix in neural networks as nonsingular. 
    
    When $d_y > d_x$, we can reformulate the leaky-ReLU network with width $N=d_y$ as $f_L(x) = \psi(W_1 x + b_1), W_1\in \mathbb{R}^{d_x\times d_y}, b\in \mathbb{R}^{d_y}$, where $\psi(\cdot)$ is a homeomorphism in dimension $d_y$. 
    \changes{
        In this sense, the universal approximation problem is related to the manifold embedding problem \cite{Hirsch1976Differential}. In particular, the Whitney's theorem indicates that $d_y=2 d_x$ is critical for embedding. Next, we will discuss the approximation ability of diffeomorphism mapping to continuous functions under the condition of whether $d_y$ is greater than $2d_x$.
    }
    
    \subsection{The particular dimensions $d_x + 1 \le d_y  \le 2d_x$ }
    
    The following lemma shows that width $N=d_y$ is not enough for $C$-UAP which implies that $w_{\min} \ge d_y+1$ when $d_y \in [d_x+1, 2 d_x]$.
    
    \begin{lemma}
        \label{th:dy=dx+1:2dx}
        If $d_y \in [d_x+1, 2 d_x]$, there exists a continuous function $f^*(x) \in C(\mathcal{K}, \mathbb{R}^{d_y})$ which can NOT be uniformly approximated by functions like $\psi(W_1 x + b_1)$ with homeomorphism maps $\psi: \mathbb{R}^{d_y} \to \mathbb{R}^{d_y}$.
    \end{lemma}
    
    \changes{
    It is easy to construct such a $f^*$ from the topology aspect by introducing regular self-intersections. For example, Whitney \cite{whitney1944Self} constructed a continuous funtion $f^* : \mathbb{R}^{d_x} \rightarrow \mathbb{R}^{2d_x}$ which maps $ (x_1,x_2,\cdots , x_{d_x})$ to $( x_1 - \frac{2x_1}{u} , x_2 ,x_3 ,\cdots ,x_{d_x},\frac{1}{u} ,\frac{x_1x_2}{u}, \frac{x_1x_3}{u},\cdots,\frac{x_1x_{d_x}}{u})$ where $u = (1+ x_1^2)(1+x_2^2)\cdots(1+x_{d_x}^2)$. The function has a regular self-intersection point $f^*(1,0,..,0)=f^*(-1,0,...,0)$, which prevents arbitrary approximation by diffeomorphisms. For the case of $d_x+1 \le d_y<2d_x$, we can leave the last $2d_x-d_y$ dimensions of $f^*$ to obtain similar results. See the appendix for the detailed proof. 
    }

    \changes{
        Here we use a simple one-dimensional example, '4'-shape curve $g(t):[0,1]\to \mathbb{R}^2$ (Figure~\ref{fig:UAP_3_dim}(a)), to intuitively show the phenomenon. In this example, we need to show that for any sufficiently small $\varepsilon>0$, the diffeomorphism map $h$, matrix $W$ and vector $b$, satisfying $\|g (t) - h (Wt+b)\|<\varepsilon$ for all $t\in[0,\tau]$ always has a self-intersection point. This is intuitively obvious by considering two continuous curves in $[-1,1]^2$, one starts from $(0,-1)$ to $(0,1)$ and another from $(-1,0)$ to $(1,0)$, which have at least one intersection point. This conclusion is so intuitive that even non-mathematics can also see it at a glance. While the proof may seem complicated because we need knowledge of topology.
    }
    
    \subsection{The case of $d_y > 2d_x$}
    
    \changes{
    Here we only consider the case of $d_y=2d_x+1$. When $d_y>2d_x+1$, the proof process can be obtained by the same method. Then, employing the lift-flow-discretization approach, we only need to show that any $f^* \in C(\mathcal{K},\mathbb{R}^{2d_x+1})$ can be approximated by functions formulated as $\psi(W x)$, where $\psi(\cdot)$ is a diffeomorphism in dimension $2d_x+1$. This is true according to Whitney's embedding theorem \cite{Hirsch1976Differential}.
    \begin{lemma}
        \label{th:dy=2dx+1}
        For any $f^* \in C(\mathcal{K},\mathbb{R}^{2d_x+1})$ and $\varepsilon>0$, there is a matrix $W \in \mathbb{R}^{(2d_x+1)\times d_x}$ and a diffeomorphism map $\Phi$ such that $\| \Phi(Wx)-f^*(x)\| < \varepsilon$ for all $x$ in $\mathcal{K}$.
    \end{lemma} 
    For any $f^* \in C(\mathcal{K},\mathbb{R}^{2d_x+1})$, Whitney embedding theorem implies that there exists an embedding $f \in C(\mathcal{K},\mathbb{R}^{2d_x+1})$ that can approximate $f^*$ well. By directly lift $\mathbb{R}^{d_x}$ as a hyperplane in $\mathbb{R}^{2d_x+1}$, $f$ can be represented as the form of $f(x)=\Phi(Wx), W\in \mathbb{R}^{(2d_x+1)\times d_x}$.
    According to the lift and flow steps, there exists a flow map $\phi^\tau \in \mathcal{C}(\mathbb{R}^{2d_x+1},\mathbb{R}^{2d_x+1})$ satisfying $\|\phi^\tau(x)-f(x)\|<\varepsilon/2$.
    According to the discretization step, there exists a leaky-ReLU network $\psi \in \mathcal{C}(\mathbb{R}^{2d_x+1},\mathbb{R}^{2d_x+1})$ that satisfies $\|\psi(x)-\phi^\tau(x)\|<\varepsilon/2$. By employing the lift-flow-discretization approach, we can arrive at the desired result. 
    }
    
\begin{figure}[htp!]
        \centering
        \includegraphics[width=8.3cm]{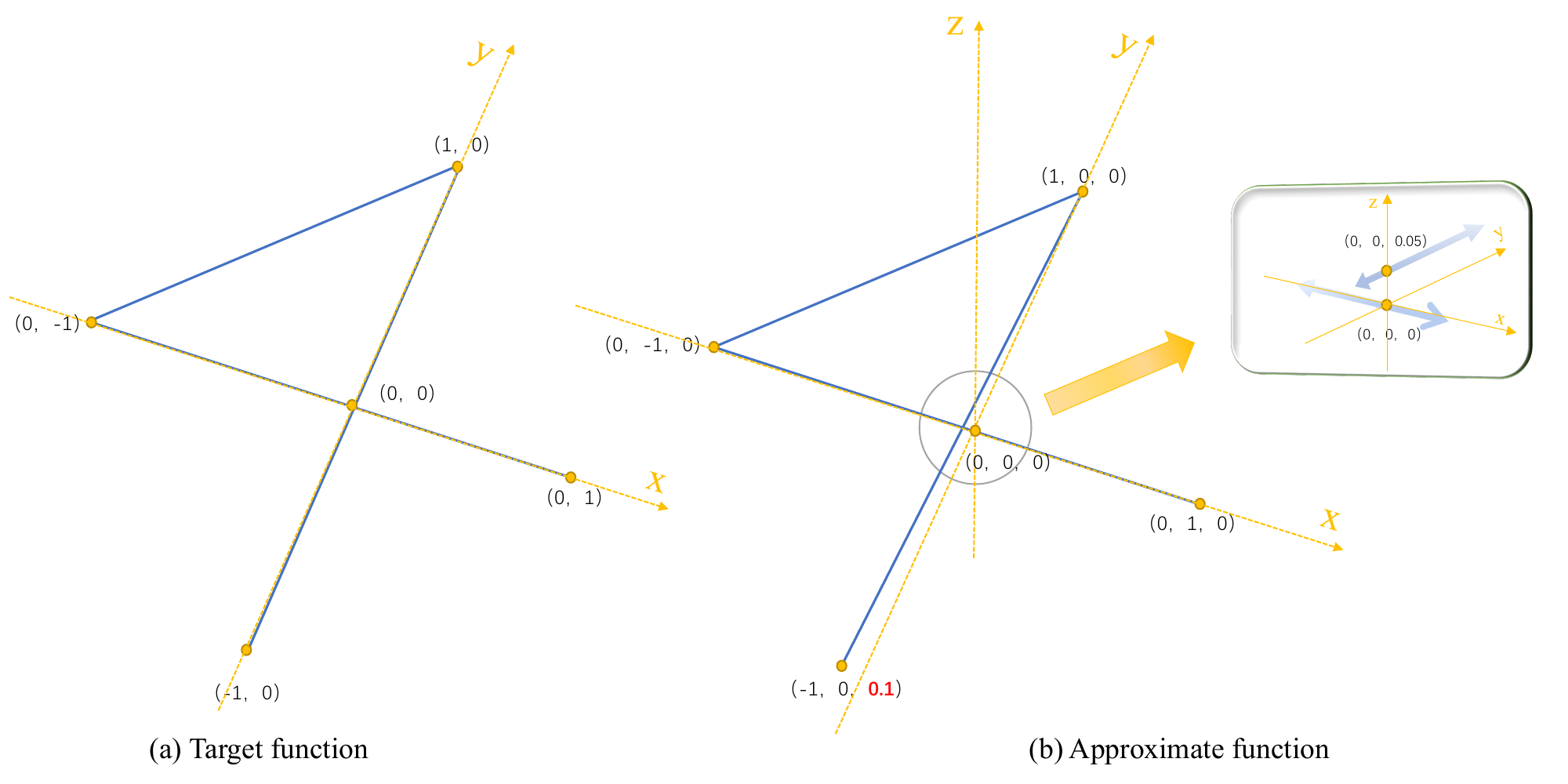}
        \caption{Example of $d_x=1$. Approximate the `4'-shape curve (a) in $\mathbb{R}^2$ by lifting it to the three-dimensional curve (b) in $\mathbb{R}^3$. }
        \label{fig:UAP_3_dim}
    \end{figure}

    To understand the result, we show an example of the case of $d_y= 2d_x+1$ in Figure \ref{fig:UAP_3_dim}(b) with $d_x=1$. It's a `4'-shape curve corresponding to a continuous function $f^*(t)$ from $[0,1] \to \mathbb{R}^3$.
    
    From Figure \ref{fig:UAP_3_dim}, we lift the four vertices of the `4'-shape curve as $(-1,0,0), (1,0,0), (0,-1,0), (0,1,0)$, and connect them in turn to form a polyline, which is our target function $f^*(t), t \in [0,1]$. Then, we construct a curve without self-intersection points by changing one of the $z$-axis coordinates of the points to $\varepsilon$ (such as $\varepsilon=0.1$), the approximation function $f$ is the curve connected by 4 vertices as $(-1,0,\varepsilon),(1,0,0),(0,-1,0),(0,1,0)$ in sequence, in Figure~\ref{fig:UAP_3_dim}(b). Now the approximation function becomes a curve $\tilde{f}(t)$ without self-intersecting points, corresponding to a homeomorphic mapping $\Phi$ in $\mathbb{R}^3$ with $\tilde{f}(t) = \Phi(w t)$ for some $w \in \mathbb{R}^{3}$. Employing the flow and discretization steps, we can approximate $\Phi$ by leaky-ReLU networks. Consequently, we can conclude that the `4'-shape curve in Figure \ref{fig:UAP_3_dim}(b) can be approximated by leaky-ReLU networks.

    \section{Discussion}
    \label{sec:discussion}

    \textbf{General Activation}. It should be noted that our `lift-flow-discretization' approach is generic for strictly monotone continuous activations. For example, our results are valid for strict monotone piecewise linear activations. We focus on leaky-ReLU networks mainly because 1) it is the simplest demo to prove our concept, and 2) the results of \cite{caponigro2011orientation} and \cite{duan2022vanilla} allow us to finish the `flow' and `discretization' steps easily.
    
    We also note that our result may not hold for ReLU networks as the ReLU function is not invertible. ReLU networks can be regarded as the limits of leaky-ReLU networks with parameter $\alpha$ tending to $0$. However, in our construction, some weights of the network are $O(1/\alpha)$, which tend to $\infty$ as $\alpha \rightarrow 0$. This suggests that the narrow ReLU and leaky-ReLU networks are different. How to rediscuss these issues under the ReLU network, and show the differences between the two more clearly, maybe a very interesting topic.

    \textbf{$L^p$-UAP and $C$-UAP}. Leaky-ReLU activation has been studied by \cite{duan2022vanilla} and \cite{cai2022achieve} to connect neural ODEs, flow maps, and the minimum width of neural networks. However, the previous results are for the UAP under the $L^p$ norm, which simplified the analysis because the diffeomorphisms are $L^p$ approximations of maps \cite{Brenier2003Approximation}. Our `lift-flow-discretization' approach can deeply connect the minimal width to the topology theory, properly lift the target function to higher dimensions, and employ facts from topology theory to further obtain sharp bounds of width for the uniform/$C$-UAP. 
    
    \textbf{Approximation rate}. Determining the number of weights or layers to achieve $\varepsilon$ approximation error is related to the approximation rate or the error-bound problems.  Estimating the error bound of all three steps in our lift-flow-discretization approach is challenging since the error in the `flow' step is hard to estimate. This may need to establish new construction tools. We leave it as future work.
    
\section*{Errata of the previous versions}

In our original ICML 2023 paper, we claimed the conclusion that the minimum width of a leaky-ReLU network required to achieve $C$-UAP for continuous functions is given by $w_{\min} = \max(d_x + 1, d_y) + 1_{d_y = d_x + 1}$. However, on September 20, 2023, we received an email from Geonho Hwang, Namjun Kim, and Sejun Park highlighting details regarding our proofs in Lemma 2.4(v1) and Lemma 4.2(v1).

Upon discussions with Mr. Lei Ziyi (who helped us formulate the proof for Lemma 4.2(v1) with $d_x=1$), he also pointed out our mistakes in generating Lemma 4.2(v1) to $d_x \ge 2$. That is $d_x+2 = 1*d_x+2 = 2*d_x+1$ only true for $d_x=1$. Mr. Ziyi also introduced us to known topological results, notably Whitney's embedding theorem, which provides not only the possibility but also the optimality for embedding. Based on these insights derived from differential topology, we have updated Lemma 4.1 and Lemma 4.2.

As for Lemma 2.4 (v1), we managed to construct a diffeomorphism $\Phi$ of $\mathbb{R}^{d+1}$, which is actually not. However, the constructed $\Phi$ is a monotone function which is still possible to be approximated by flow maps. This is the main claim of a recent work of Tabuada et al. \cite{Tabuada2020Universal}. Employing the Corollary 5.2 of \cite{Tabuada2020Universal}, we updated the manuscript to v2 on October 8, 2023. 

\textbf{Claim:}
\emph{
    For any compact set $\mathcal{K}\subset\mathbb{R}^d$, continuous function $f^* \in C(\mathcal{K},\mathbb{R}^{d})$, and $\varepsilon>0$, there exists a flow map $\phi^\tau \in C(\mathbb{R}^{d+1},\mathbb{R}^{d+1})$ and two linear maps, $\alpha$ and $\beta$, such that $\|f^*(x)-\beta\circ\phi^\tau \circ\alpha(x)\|<\varepsilon$ for all $x$ in $\mathcal{K}$.
}

Based on this claim, the minimum width can be obtained as $w_{\min} = \max(d_x+1,d_y)+1_{d_y=d_x+1:2d_x}$. However, Tabuada et al. got this claim from the control theory, and when we try to understand it from topology theory, we find a monotonic function $f^*$ that may not be approximated by diffeomorphism (see Appendix \ref{sec:Counterexample}). We had a discussion with the authors of \cite{Tabuada2020Universal} about this counterexample $f^*$ on October 3, 2023. When we go deep into the proof in \cite{Tabuada2020Universal}, we find some potential assumptions are made in their work, which is confirmed by the authors of \cite{Tabuada2020Universal} on January 12, 2024. As a result, determining the values of function $\Delta(d_x,d_y)$ becomes open. We left it to experts in topology.

\begin{table*}[htpb!]
    \centering
    \caption{Additional results on the minimum width of leaky-ReLU networks}
    \setlength{\tabcolsep}{1mm}{
        \begin{tabular}{cccc}
            \toprule[1.5pt]
            References & Functions & Minimum width\\
            \hline
            \cite{Hanin2018Approximating} & $\mathcal{C}(\mathcal{K},\mathbb{R}^{d_y})$ & $d_x+1 \le w_{\min} \le d_x+d_y$\\
            \hline
            \cite{hwang2023minimum} & $\mathcal{C}(\mathcal{K},\mathbb{R}^{d_y})$ & $w_{\min} \le \max(2d_x + 1,d_y)$\\
            \hline
            & $\mathcal{C}([0,1]^2,\mathbb{R}^{2})$ & 4\\
            \hline
            \cite{kim2023minimum} & $\mathcal{C}([0,1]^{d_x},\mathbb{R}^{d_y})$  &$w_{\min}\ge d_y+1_{d_x<d_y\le 2d_x}$\\
            \bottomrule[1.5pt]
        \end{tabular}\\
    }\label{tab:Errata}
\end{table*}

Although the results in the current version are not as strong as the original ICML version, it still reveals new ideas for understanding and solving the minimum width for UAP. It can prove many existing conclusions concisely and intuitively and give a unified explanation from a topological perspective. To illustrate the merit of this argument, we recall the result of \cite{Hanin2018Approximating} which proves that the minimum width $w_{\min}$ to implement $C$-UAP under the ReLU network satisfies $d_x+1 \le w_{\min} \le d_x+d_y$. Since the leaky-ReLU network can approximate the ReLU network, this conclusion is also true for the leaky-ReLU network (see Table~\ref{tab:Errata}). We write it as the following corollary which can be concisely proved in the same way as Lemma \ref{th:w_upper_bound} from a topological perspective (see Appendix \ref{sec:Proof_Corollary5.1}).

\begin{corollary}
    \label{minimum}
    For any continuous function $f^* \in C(\mathcal{K},\mathbb{R}^{d_y})$ on compact domain $\mathcal{K} \subset \mathbb{R}^{d_x}$, the minimum width $w_{\min}$ of leaky-ReLU neural networks having C-UAP satisfies  $d_x+1 \le w_{\min} \le d_x +d_y$.
\end{corollary}

During the revision period of this paper, there are some parallel progresses in the direction of determining the minimal width of leaky-ReLU networks which is also listed in Table~\ref{tab:Errata}. In particular, when we were ready to update this version, we found that \cite{hwang2023minimum} already showed a new result that $w_{\min} = 4$ if $d_x=d_y=2$. It suggests that the upper bound in \cite{Hanin2018Approximating} might be tight for many cases.

\section*{Acknowledgments}

\changes{
We are grateful to Mr. Ziyi Lei for the helpful discussion on the proof of lemmas in Section \ref{sec4}. We are also grateful to the anonymous reviewers for their useful comments and suggestions. In addition, we thank Geonho Hwang,
Sejun Park, and Namjun Kim, for locating our mistakes in Lemma \ref{th:w_upper_bound} and Lemma \ref{th:dy=2dx+1} of the first version. 
We also thank Paulo Tabuada and Bahman Gharesifard for discussing their results. This work was supported by the National Natural Science Foundation of China (Grant No. 12201053) and the National Natural Science Foundation of China (Grant No. 11871105).
}

\bibliography{refs.bib}

\bibliographystyle{icml2023}

\newpage
\appendix
\onecolumn

\section{Notations}

\textbf{Flow map} A flow on a set $\mathbb{X}$ is a group action of the additive group of real numbers on $\mathbb{X}$. More explicitly, a flow is a mapping
$\phi: \mathbb{X} \times \mathbb{R} \to \mathbb{X} $ such that for all $x \in \mathbb{X}$ and  $s,t \in \mathbb{R}$,
$\phi(x,0)=x,\phi(\phi(x,t),s)=\phi(x,s+t)$. If we fix $t=\tau$, we gain a flow map $\phi^\tau:\mathbb{X}\to \mathbb{X}$.

\textbf{Diffeomorphism} Given two manifolds $M$ and $N$, a differentiable map $f: M \rightarrow N$ is called a diffeomorphism if it is a bijection and its inverse $f^{-1}: N \rightarrow M $ is differentiable as well.

\section{Proofs}

\subsection{Proof of Lemma \ref{2.3}}
\begin{proof}
 To illustrate this lemma, we construct two linear maps $\alpha : \mathbb{R}^{d_x} \rightarrow  \mathbb{R}^{N}, (x_1,\cdots ,x_{d_x})  \rightarrow (x_1,\cdots ,x_{d_x},0,\cdots,0)$ and $\beta :\mathbb{R}^{N} \rightarrow  \mathbb{R}^{d_y}, (x_1,\cdots ,x_{N})  \rightarrow (x_1,\cdots ,x_{d_y})$.
    
     For any $\varepsilon>0$, there is a leaky-ReLU network (\ref{eq:FNN}) $$f(x) =  W_{L+1} \sigma (W_L ( \cdots \sigma ( W_1 x + b_1)+\cdots) + b_L) + b_{L+1} = \tilde \beta \circ \tilde \Phi \circ \tilde \alpha$$ such that $\|f(x) - f^*(x)\| \le \frac{\varepsilon}{2}$ for all $x$ in $\mathcal{K}$, where $\tilde \Phi : \mathbb{R}^{N} \rightarrow \mathbb{R}^{N}$ is a reversible mapping. Since leaky-ReLU is piecewise linear, the $\tilde \Phi$ here is not a diffeomorphism. Note that a singular matrix can be approximated by nonsingular matrices, $W_i$ can be well approximated by some nonsingular matrix $\hat W_i$. Just like the ReLU function can be smoothly approximated by the softplus function, $\sigma(x)$ can be well approximated by its smooth version $\hat \sigma(x) = (\ln(e^{\lambda x} + 1) - \alpha \cdot \ln(e^{-\lambda x}+1))/\lambda$ where $\lambda >0$ is large enough. Therefore, we can construct a neural network to approximate the leaky-ReLU network $f$ like  $$\hat{f}(x) = \hat W_{L+1} \hat \sigma (\hat W_L ( \cdots \hat \sigma (\hat W_1 x + \hat b_1)+\cdots) + \hat b_L) + \hat b_{L+1} =  \beta \circ  \Phi \circ \alpha$$ such that $\|\hat{f}(x) - f(x)\| \le \frac{\varepsilon}{2},$ where $\Phi: \mathbb{R}^{N} \rightarrow \mathbb{R}^{N}$ is a diffeomorphism.
    So we can get a diffeomorphism $\Phi$ of $\mathbb{R}^{N}$ satisfying $$\|f^*(x) - \beta \circ \Phi \circ \alpha (x)\| \le \|\hat{f}(x) - f(x)\| + \|f(x) - f^*(x)\|= \varepsilon.$$
    This means that $f^*$ can be approximated by a diffeomorphism, that is, $w_{\min} \ge \max(d_x,d_y)+\Delta (d_x , d_y)$.
\end{proof}

\subsection{Proof of Lemma \ref{th:appproximate_OP_diff}}

\begin{proof}
    It is a corollary of Theorem 6
    in 
    \cite{caponigro2011orientation}. 
    Here, we only provide the main ideas.
    Let $\mathcal{M} \subset \mathbb{R}^N$ be a compact connected manifold and $\{f_1,\cdots,f_n\}$ be a bracket-generating family of vector fields. \cite{caponigro2011orientation} shows that 
     for any OP diffeomorphism $P \in \text{Diff}_0(\mathcal{M})$ and $\varepsilon>0$,
     there exist $n$ time-varying feedback controls, $u_j(x,t)$, 
     which are piecewise constant with respect to $t$, such that $P$ 
     can be represented by the flow map $\tilde{\phi}^\tau$ of the ODE 
     $\dot x(t) = \sum\limits_{j=1}^n u_j(x,t)f_j(x), t\in(0,\tau)$, which 
     means $\|\tilde{\phi}^\tau(x)-\Phi(x)\|<\varepsilon$. 
     
    Then, we find a neural ODE to approximate $\tilde{\phi}^\tau$.
    For the approximation of $\tilde{\phi}^\tau$, we need to approximate each
    $u_j(x,t)$, which can be done by polynomial, trigonometric 
    polynomials or neural networks. The neural ODE (\ref{eq:tanh_ODE}) is 
    such an example that takes $\mathcal{M} \supset \Omega$, $n=N$,
    $f_j = e_j$ as the axis vectors, and according to the UAP of 
    neural networks, we have
    $u_j(x,t) \approx \sum_{i=1}^M
    a_{i}^{(j)}(t) \tanh(w_i(t) \cdot x+b_i(t)) $ 
    where $a_{i}^{(j)}$ is the $j$-th coordinate of $a_i$. In this case, ${\phi}^\tau$,
    the flow map of Eq.~(\ref{eq:tanh_ODE}) satisfies $\|\tilde{\phi}^\tau(x)-\phi^\tau(x)\|<\varepsilon,x\in \mathbb{R}^N$.
\end{proof}

\subsection{Proof of Lemma \ref{th:appproximate_flowmap}}

\begin{proof}
    The main idea is to solve the ODE (\ref{eq:tanh_ODE}) by a splitting method and then approximate each split step by leaky-ReLU networks. Consider the following splitting for $v$ in (\ref{eq:tanh_ODE}):
    $ v(x,t) 
    \equiv
    \sum_{i=1}^N \sum_{j=1}^d 
    v_{ij}(x,t) 
    {e}_j$ with $v_{ij}(x,t) = a_{i}^{(j)}(t) \tanh(w_i(t) \cdot x+b_i(t)) \in \mathbb{R}$.
    Then, the flow map can be approximated by an iteration with time step $\Delta t = \tau/n$, $n \in \mathbb{Z}^+$,
    \begin{align*}
        \phi^{\tau}(x_0) \approx  x_n
        &= T_n (x_{n-1}) = T_n \circ \cdots \circ T_1 (x_0),\\
        &=
        T_n^{(N,d)} \circ \dots \circ T_n^{(1,2)} \circ T_n^{(1,1)} 
        \circ
        \cdots \circ
    T_1^{(N,d)} \circ \cdots \circ T_1^{(1,2)} \circ T_1^{(1,1)} 
        (x_0).
    \end{align*}
    where the $k$-th iteration is $x_{k+1} = T_k x_k = T_k^{(N,d)} \circ \dots \circ T_k^{(1,2)} \circ T_k^{(1,1)} (x_k)$.
    The map $T_k^{(i,j)}: x \to y$ in each split step is:
    \begin{align*}
        T_k^{(i,j)}:
        \left\{
        \begin{aligned} 
        & y^{(l)} = x^{(l)} , l = 1,2,..,j-1,j+1,...,d,  \\
        & y^{(j)} = x^{(j)} + \Delta t v_{ij}(x,k\Delta t).
        \end{aligned}
        \right.
    \end{align*}
    Here, the superscript in $x^{(l)}$ indicates the $l$-th coordinate of $x$. \cite{duan2022vanilla} constructed leaky-ReLU networks with width $N$ to approximate each map $T_k^{(i,j)}$ and we finished the proof. 

\end{proof}

\subsection{Proof of Corollary \ref{th:LFD_approach}}

\begin{proof}
    This corollary can be directly derived from the `lift-flow-discretization' steps.
\end{proof}

\subsection{Proof of Lemma \ref{th:dy=dx+1:2dx}}

\begin{proof}
    Following \cite{whitney1944Self}, we consider the following function $f^* : [-2, 2]^{d_x} \rightarrow \mathbb{R}^{d_y}$,$ d_y \in [d_x +1 , 2 d_x]$,
     \begin{align}
        f^*(x): = \Big( x_1 - \frac{2x_1}{u} , x_2 ,x_3 ,\cdots ,x_{d_x},\frac{1}{u} ,\frac{x_1x_2}{u}, \frac{x_1x_3}{u},\cdots,\frac{x_1x_{d_y-d_x}}{u} \Big).
    \end{align}
In particular, when $d_y=d_x + 1$ ,$ f^*(x) := ( x_1 - \frac{2x_1}{u} , x_2 ,x_3 ,\cdots ,x_{d_x},\frac{1}{u})$,
where $u = (1+ x_1^2)(1+x_2^2)\cdots(1+x_{d_x}^2)$.
It's easy to find that $f^*$ has a self-intersection $P_0=f^*(1,0,\cdots,0) = f^*(-1,0,\cdots,0)$. Then we show that it is a regular intersection. The matrix of partial derivative transposed at $(\pm 1, 0,... , 0)$ is $M^{\pm}$,
\begin{align}
M^{\pm} = 
    \left (\begin{array}{ccccccccc:cccccc}
    1 & 0 & 0& \ldots & 0 & 0 &\ldots & 0 & 0& \mp \frac{1}{2} & 0  & 0& \ldots & 0 & 0 \\
    0 & 1 & 0& \ldots & 0 & 0 &\ldots & 0 & 0& 0 & \pm \frac{1}{2}  & 0& \ldots & 0 & 0 \\
    0 & 0 & 1& \ldots & 0 & 0 &\ldots & 0 & 0& 0 & 0 & \pm \frac{1}{2} & \ldots & 0 & 0 \\
    \vdots & \vdots & \vdots & \ddots & \vdots & \vdots & \ddots & \vdots & \vdots & \vdots & \vdots & \vdots & \ddots & \vdots & \vdots \\
    0 & 0 & 0& \ldots &1 & 0 &\ldots & 0 & 0 & 0 & 0  & 0& \ldots & 0 & \pm \frac{1}{2} \\
    \hdashline
    0 & 0 & 0& \ldots &0 & 1 &\ldots & 0 & 0 & 0 & 0  & 0& \ldots & 0 & 0 \\
    \vdots & \vdots & \vdots & \ddots & \vdots & \vdots & \ddots & \vdots & \vdots & \vdots & \vdots & \vdots & \ddots & \vdots & \vdots \\
    0 & 0 & 0& \ldots & 0 & 0 &\ldots & 1 & 0 & 0 & 0  & 0& \ldots & 0 & 0 \\
    0 & 0 & 0& \ldots & 0 & 0 &\ldots & 0 & 1 & 0 & 0  & 0& \ldots & 0 & 0 \\
    \end{array}\right)_{d_x \times d_y}.
\end{align}
The elements of $M^{\pm}_{ij}$ are  $\frac{\partial f^*_j}{\partial x_i}(\pm 1, 0, ... 0)$. Select all rows of $M^+$ and the first $d_y-d_x$ rows of $M^-$ to form the following matrix $M \in \mathbb{R}^{d_y \times d_y}$,
\begin{align}
    M = \left (\begin{array}{c}
        M^+_{1:d_x,:} \\
        M^-_{1:d_y-d_x,:}
        \end{array}\right),
\end{align}
which is nonsingular because its determinant is $\det(M)=1$. Correspondingly, define two manifolds, $ \mathcal{M}_1$ and $ \mathcal{M}_2$, as below,
\begin{align}
    \mathcal{M}_1 &= \{ f^*(x_1, ..., x_{d_x}) : (x_1, ..., x_{d_x}) \in U_\delta ( 1 , 0 , ... , 0 )\},\\
    \mathcal{M}_2 &= \{ f^*(x_1, ..., x_{d_x}) : (x_1, ..., x_{d_x}) \in U_\delta (-1 , 0 , ... , 0 ), x_{d_y-d_x+1}=...=x_{d_x}=0\},
\end{align}
which has one intersection point $P_0$, where $U_\delta$ means a small neighborhood. Since the tangent spaces of $ \mathcal{M}_1$ and $ \mathcal{M}_2$ at $P_0$ are spanned by the corresponding rows of $M$, the nonsingularity of $M$ implies $P_0$ is a transverse intersection which can not be eliminated by small perturbations (this result is standard, see Lemma \ref{intersect} for example). In other words, no diffeomorphism $\psi$ can arbitrarily approximate $f^*$, which finishes the proof.
\end{proof}

\begin{lemma}
    \label{intersect}
    Two smooth manifolds that transversely intersect cannot eliminate the intersection point by small perturbations.
\end{lemma}

\begin{proof}
    Without loss of the generality, we assume the manifolds are hyperplanes, $F^{r}$ and $F^{s}$, in the following form,
    \begin{align}
        F^{r} &= \{x \in \mathbb{R}^{r+s}| x_1 = \cdots = x_{s}=0\},\\
        F^{s} &= \{x \in \mathbb{R}^{r+s}| x_{s+1} = \cdots = x_{s+r}=0\}.
    \end{align}

We can define a map $f: \mathbb{R}^{r+s} \to \mathbb{R}^{r+s} , x \to (x_1,\cdots , x_{s},-x_{s+1},\cdots,-x_{r+s})$. As the domation of $f$, $\mathbb{R}^{r+s}$ is actually $F^r \times F^s$, the value of $f$ is actually the minus of a point in $F^{r} \subseteq \mathbb{R}^{r+s}$ and another one in $F^{s} \subseteq \mathbb{R}^{r+s}$. This implies that $f$ is a one-to-one mapping, and $f^{-1}(0)$ is the set of intersection points of the two hyperplanes.

If there is a pair of diffeomorphism approximation of $F^{r}, F^{s}$ in $\mathbb{R}^{r+s}$ without intersection, the map$f$ has a diffeomorphism approximation $\widetilde{f}: \mathbb{R}^{r+s} \to \mathbb{R}^{r+s}$, s.t. $||\widetilde{f}-f||<\epsilon$ for $\epsilon$ small enough, and no point is mapped to 0.

Because $f(x) \to \infty,\widetilde{f}(x) \to \infty$ when $x \to \infty$, we can extend $\widetilde{f}(x)$ to $\mathbb{S}^{r+s} := \mathbb{R}^{r+s} \cup \{\infty\}.$ That is, $\bar{f}: \mathbb{S}^{r+s} \to \mathbb{S}^{r+s}, \bar{f}|_{\mathbb{R}^{r+s}}= \widetilde{f}$, and $\bar{f}^{-1}\{0\}= \emptyset.$ In addition, $\bar{f}$ is a diffeomorphism from $\mathbb{S}^{r+s}$ to $\bar{f}(\mathbb{S}^{r+s})$ as $\widetilde{f}$ is. Hence $\bar{f}$ is an embedding but not surjective, which is impossible. 
The reason is that $\mathbb{S}^{r+s}$ is compact and has no boundary which implie that $\bar{f}(\mathbb{S}^{r+s})$ is closed in $\mathbb{S}^{r+s}$. In addition, $\bar{f}(\mathbb{S}^{r+s})$ is open in $\mathbb{S}^{r+s}$ as $\bar{f}$ is a diffeomorphism. Therefore $\bar{f}(\mathbb{S}^{r+s})$ is both open and closed in $\mathbb{S}^{r+s}$, and hence must be $\mathbb{S}^{r+s}$ or $\emptyset$, which is contradictory to the argument that $\bar{f}$ is an embedding but not surjective.

So we show the map $f$ cannot be approximated by diffeomorphisms of the same dimension without intersection.
\end{proof}

\subsection{Proof of Lemma \ref{th:dy=2dx+1}}

\begin{proof}

According to Whitney's embedding theorem (see \cite{Hirsch1976Differential} for example), $f^*$ can be approximated by a smooth embedding $f$. Let $W$ be a linear map which lifts $\mathcal{K}$ to dimension $d_y$, then $f$ can be represented as $\Phi(Wx)$ where $\Phi$ is a diffeomorphism and $\Phi(Wx)$ aapproximates $f^*(x)$.
\end{proof}

\subsection{Proof of Theorem \ref{th:main_min_width}}

\begin{proof}
The proof is divided into two parts. On the one hand, from the discussion of Lemma \ref{2.3}, we can get a lower bound $w_{\min} \ge \max(d_x,d_y)+\Delta (d_x , d_y)$.
On the other hand, Corollary \ref{th:LFD_approach} shows that this is also an upper bound $w_{\min} \le \max(d_x,d_y)+\Delta (d_x , d_y)$. The theorem is proved.

\end{proof}

\subsection{Proof of Corollary \ref{minimum}}
\label{sec:Proof_Corollary5.1}
     As we all know, every continuous function defined on a closed interval can be uniformly approximated as closely as desired by a polynomial function according to the Weierstrass approximation theorem. 
    Without loss of generality, we assume $f^*$ to be a polynomial function.
    Taking two linear maps $\alpha : \mathbb{R}^{d_x} \rightarrow  \mathbb{R}^{d_x +d_y}, (x_1,\cdots ,x_{d_x})  \rightarrow (x_1,\cdots ,x_{d_x},0,\cdots,0)$ and $\beta :\mathbb{R}^{d_x +d_y} \rightarrow  \mathbb{R}^{d_y}, (x_1,\cdots ,x_{d_x +d_y})  \rightarrow (x_{d_y+1},\cdots ,x_{d_x +d_y})$, we can construct the mapping $\Phi : \mathbb{R}^{d_x + d_y} \rightarrow  \mathbb{R}^{d_x + d_y}, (x_1,\cdots ,x_{d_x+d_y})  \rightarrow (x_1,\cdots,x_{d_x},y_1 + x_{d_x +1},\cdots ,y_{d_y} + x_{d_x +d_y})$  to satisfy $f^* = \beta \circ \Phi \circ \alpha (x)$. Since $f^*$  is a continuous function, $\Phi$ is obviously a continuous one-to-one mapping. The Jacobian matrix
    $$
    \nabla \Phi  = \begin{pmatrix} I_{d_x} & M \\ O & I_{d_y} \end{pmatrix}
    $$
    is an upper triangular matrix whose determinant is one, where $I_{d_x}$ and $I_{d_y}$ are identity matrices, and $O $ is a null matrix.
    It implies that $\Phi$ is a diffeomorphism. According to the Corollary \ref{th:LFD_approach}, , leaky-ReLU network $f_L \in \mathcal{N}_{d_x +d_y}(\sigma)$ that satisfies the hypothesis exists, $w_{\min} \le d_x + d_y$. On the other hand, Lemma \ref{th:w_upper_bound} shows that $w_{\min} \ge d_x+1$. Combining the above two parts, the corollary is proved.

\section{Counterexample of monotonic functions}
\label{sec:Counterexample}

For a two-dimensional continuous function $f^* : (x,y) \rightarrow (x^2,y^2), (x,y) \in [-1,1]^2$, $\nabla f^* = \begin{pmatrix} 2x & 0 \\ 0 & 2y \end{pmatrix}$. Therefore, we can set $k=3$ to construct a three-dimensional monotonic function $\Phi$,
$$\Phi :
\begin{pmatrix} x \\y \\z \end{pmatrix}  
\mapsto 
\begin{pmatrix} x^2 + 3x+3y \\y^2 + 3x +3y \\z  
\end{pmatrix}, 
\begin{pmatrix} x \\y \\z 
\end{pmatrix} \in [-1,1]^3
,\quad
\nabla \Phi = 
\begin{pmatrix} 2x + 3 & 3 & 0 \\ 3 & 2y+3 & 0 \\ 0 & 0 & 1 
\end{pmatrix}.
$$
We select two points $P_1 :(\frac{1}{2} , -\frac{1}{2} ,0), P_2 : (-\frac{1}{2},\frac{1}{2},0)$, and there is a self-intersection point $P = \Phi(\frac{1}{2} , -\frac{1}{2} ,0) = \Phi(-\frac{1}{2} , \frac{1}{2} ,0)$. The matrices of partial derivative transposed are:
$$ 
\nabla \Phi |_{P_1} = \begin{pmatrix} 4 & 3 & 0 \\ 3 & 2 & 0 \\ 0 & 0 & 1 \end{pmatrix} ,\quad
\nabla \Phi |_{P_2} = \begin{pmatrix} 2 & 3 & 0 \\ 3 & 4 & 0 \\ 0 & 0 & 1 \end{pmatrix} ,
$$
which means the self-intersection point $P$ can't be eliminated by perturbation. So, the monotonic function $\Phi$ can't be approximated by diffeomorphisms, which conflicts with the conclusion in \cite{Tabuada2020Universal}.

\end{document}